\documentclass[lettersize,journal]{IEEEtran}
\usepackage{amsmath,amsfonts}
\usepackage{algorithm}
\usepackage{algpseudocode}  
\usepackage{array}
\usepackage[caption=false,font=normalsize,labelfont=sf,textfont=sf]{subfig}
\usepackage{textcomp}
\usepackage{stfloats}
\usepackage{url}
\usepackage{verbatim}
\usepackage{graphicx}
\usepackage{cite}
\usepackage{color,xcolor}
\usepackage{booktabs}       
\usepackage{hyperref}
\usepackage{multirow}
\usepackage{graphicx}
\usepackage{balance}   
\usepackage{flushend}
\usepackage{amssymb}   
\usepackage{stfloats}

\begin{document}

\title{Skypilot: Fine-Tuning LLM with Physical Grounding for AAV Coverage Search}

\author{{ Zhongkai Chen$^{\dagger}$, Yihao Sun$^{\dagger}$,  Chao Yan, Han Zhou, Xiaojia Xiang and Jie Jiang}
\thanks{$^\dagger$Zhongkai Chen and Yihao Sun contributed equally to this work.}
\thanks{This work was supported in part by the National Natural Science Foundation of China under Grant 62403240 and 62533012; in part by the Natural Science Foundation of Jiangsu Province under Grant BK20241396; in part by the Postgraduate Scientific Research Innovation Project of Hunan Province under Grant CX20240114.  \textit{(Corresponding author: Chao Yan.)}}
	\thanks{Zhongkai Chen, Yihao Sun, Han Zhou and Xiaojia Xiang are with College of Intelligence Science and Technology,
	National University of Defense Technology, Changsha 410073, Hunan,
	China (e-mail: \href{mailto:chenzhongkai@nudt.edu.cn}{chenzhongkai@nudt.edu.cn}, \href{mailto:sunyihao@nudt.edu.cn}{sunyihao@nudt.edu.cn}, \href{mailto:zhouhan@nudt.edu.cn}{zhouhan@nudt.edu.cn} and \href{mailto:xiangxiaojia@nudt.edu.cn}{xiangxiaojia@nudt.edu.cn}).}%
	\thanks{Chao Yan is with College of Automation Engineering, Nanjing University of Aeronautics and Astronautics, Nanjing 211106, Jiangsu, China (e-mail: \href{mailto:yanchao@nuaa.edu.cn}{yanchao@nuaa.edu.cn}).}%
\thanks{Jie Jiang is with China Academy of Launch Vehicle Technology, Beijing 100076, China (e-mail: \href{mailto:yhztjs@163.com}{yhztjs@163.com}).}

}

\markboth{Journal of \LaTeX\ Class Files,~Vol.~14, No.~8, August~2021}%
{Shell \MakeLowercase{\textit{et al.}}: A Sample Article Using IEEEtran.cls for IEEE Journals}


\maketitle

\begin{abstract}
		Autonomous aerial vehicles (AAVs) have played a pivotal role in coverage operations and search missions. Recent advances in large language models (LLMs) offer promising opportunities to augment AAV intelligence. These advances help address complex challenges like area coverage optimization, dynamic path planning, and adaptive decision-making.
However, the absence of physical grounding in LLMs leads to hallucination and reproducibility problems in   spatial reasoning and  decision-making.
To tackle these issues,  we present Skypilot, an LLM-enhanced two-stage framework that grounds language models in physical reality by integrating monte carlo tree search (MCTS). In the first stage, we introduce  a diversified action space that encompasses generate, regenerate, fine-tune, and evaluate operations, coupled with physics-informed reward functions to ensure trajectory feasibility. In the second stage, we fine-tune Qwen3-4B on 23,000 MCTS-generated samples, achieving substantial inference acceleration while maintaining solution quality. Extensive
numerical simulations and real-world flight experiments validate the efficiency and superiority of our proposed approach. Detailed information and experimental results are accessible at \url{https://sky-pilot.top}.

\end{abstract}


	\section{INTRODUCTION}
Autonomous aerial vehicles (AAVs) have emerged as transformative platforms for executing coverage search missions across diverse application domains. These applications range from rapidly surveying disaster-stricken areas in search and rescue operations~\cite{10637661, 10433718, sun2022multi} to monitoring vast ecosystems for environmental changes~\cite{9916071} and inspecting critical infrastructure~\cite{ZHENG2025113556}. In these diverse scenarios, AAVs  provide distinct advantages: extended operational range, safe access to hazardous locations, and real-time data collection for time-sensitive decisions. To maximize these capabilities, coverage search methods guide AAVs in systematically exploring target areas while optimizing resource use. 
This involves  generating coverage-optimal trajectories, navigating around  environmental obstacles, and seamlessly switching between search patterns based on real-time mission feedback.

Traditional coverage path planning approaches for AAVs predominantly rely on geometric decomposition methods~\cite{feng2024fc}, graph-based algorithms~\cite{dong2024fast}, and optimization techniques such as genetic algorithms~\cite{KONG2025102821} or particle swarm optimization~\cite{PHUNG2021107376}. While these methods have demonstrated success in structured environments with well-defined objectives, they suffer from several critical limitations. First, most traditional algorithms require precise prior knowledge of the environment and depend on carefully crafted heuristics that lack robustness to environmental variations~\cite{zhang2024falcon}. Second, these approaches typically optimize for a single objective (e.g., minimizing path length or maximizing coverage), struggling to balance multiple competing constraints such as no-fly zone avoidance, and mission-specific requirements~\cite{datsko2024energy,10175546}.  Consequently, traditional methods may achieve   coverage patterns that are mathematically optimal but practically unsuitable for specific operational contexts~\cite{foster2025efficient}.

With the advancement of artificial intelligence, learning-based methods have garnered widespread attention. These methods include imitation learning from expert demonstrations~\cite{sui2020formation}, reinforcement learning through trial and error~\cite{10779446}, and supervised learning from labeled datasets~\cite{scarciglia2025map}. Compared to traditional methods, they offer superior adaptability to dynamic environments and handle complex state spaces without explicit mathematical modeling.  However, they also face their own set of challenges. Imitation learning methods suffer from distribution shift~\cite{lu2024you}. Reinforcement learning approaches face issues with sample inefficiency and often struggle with sim-to-real transfer~\cite{11017653}. Supervised learning methods are inherently limited by the quality and diversity of training data, often failing to generalize to scenarios beyond their training distribution~\cite{yue2024semantic}. 

In addition, both traditional  and learning-based methods lack the ability to  incorporate real-time contextual information or high-level semantic constraints, limiting their flexibility for mission-specific requirements~\cite{wieczorek2024framework}.
The recent proliferation of large language models (LLMs) presents a new paradigm for robotic control systems~\cite{firoozi2025foundation,10643253}. LLMs possess remarkable capabilities in understanding complex natural language instructions, reasoning about spatial relationships, and generating contextually appropriate control strategies~\cite{pan2025chainofaction,10933798}. Unlike traditional methods, LLM-based approaches can seamlessly integrate high-level mission objectives, operational constraints, and domain expertise expressed in natural language, enabling more intuitive human-robot collaboration~\cite{kannan2024smart,mandi2024roco}. Moreover, the emergent reasoning capabilities of LLMs allow them to adapt coverage strategies based on mission context, prioritize areas of interest based on semantic understanding, and generate explanations for their planning decisions~\cite{hazra2024saycanpay, zhang2025grounded}. However, the direct application of LLMs to path planning still faces critical challenges related to reproducibility and hallucination, which may compromise the reliability required for autonomous flight operations.

A promising solution to address these limitations is the integration of conventional sampling-based algorithms~\cite{hao2023reasoning, qi2025mutual} with LLMs. These algorithms provide principled frameworks for systematic exploration and exploitation, enabling LLMs to evaluate multiple trajectory candidates through structured sampling and evaluation processes~\cite{zhang2024accessing}. By grounding decisions in simulation-based feedback,  these algorithms not only validates and refines LLM outputs but also effectively reduces hallucination~\cite{tian2024toward}. Additionally, sampling-based methods can effectively balance global exploration with local optimization through various heuristic mechanisms~\cite{kocsis2006bandit}. However, such algorithms typically require numerous LLM inference calls per iteration, creating significant computational overhead for real-time planning. The computational cost also scales poorly with environmental complexity and planning horizons, making online deployment impractical for resource-constrained aerial platforms.

In this work, we propose Skypilot, a two-stage LLM-based coverage search method for AAV exploration missions. In the first stage, we develop a novel  monte carlo tree search (MCTS) architecture that leverages LLMs to generate and evaluate coverage trajectories while satisfying complex operational constraints. Specifically, we design a diversified action space encompassing generation, regeneration, fine-tuning, and evaluation operations. Additionally, we present a physics-informed reward function that guides LLMs to produce feasible coverage paths while respecting both physical constraints and human instructions. In the second stage, to alleviate the computational burden of online MCTS inference, we construct a high-quality dataset of 23,000 trajectory samples through systematic MCTS rollouts. Then we introduce a novel fine-tuning strategy that employs full-parameter optimization, diverging from conventional low-rank adaptation (LoRA) approaches~\cite{hu2022lora} for comprehensive model adaptation.  Comprehensive empirical evaluations demonstrate that Skypilot significantly outperforms baseline methods in coverage efficiency, constraint satisfaction, and computational scalability. The main contributions of this work can be summarized as follows:

1)
We propose a novel MCTS framework   tailored for LLM-based coverage planning to systematically explore the solution space while ensuring generated trajectories satisfy both coverage objectives and operational constraints.

2) We introduce an efficient fine-tuning strategy using the constructed dataset that significantly reduces model inference time while preserving exceptional planning performance for real-time coverage.

3) We validate the effectiveness and applicability of Skypilot through comprehensive indoor and outdoor experiments with real AAV platforms. In these experiments, we construct an interactive ground station for AAV control, demonstrating significant mission performance improvements.

The remainder of this paper is organized as follows. Section \ref{pf} provides the problem formulation for LLM-enhanced AAV coverage  mission.  Section \ref{me} presents our proposed Skypilot framework. Section \ref{exp} introduces simulation and experimental results. Finally, Section \ref{con} concludes this work.

\begin{figure}[t]
	\centering
	\includegraphics[width=0.99\linewidth]{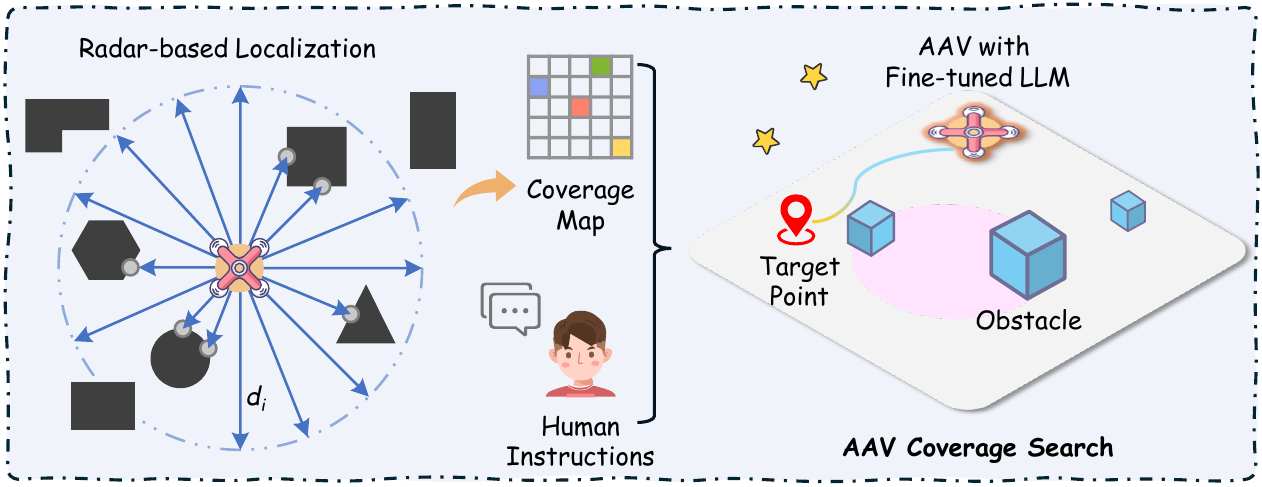}
	
	\caption{
		Illustration of the AAV coverage search task. Coverage maps from radar-based environmental perception and human instructions are jointly processed by a fine-tuned LLM to guide AAV coverage search operations.
	}
	\label{fig:scenario}
\end{figure}

\begin{figure*}[t]
	\centering
	\includegraphics[width=0.99\linewidth]{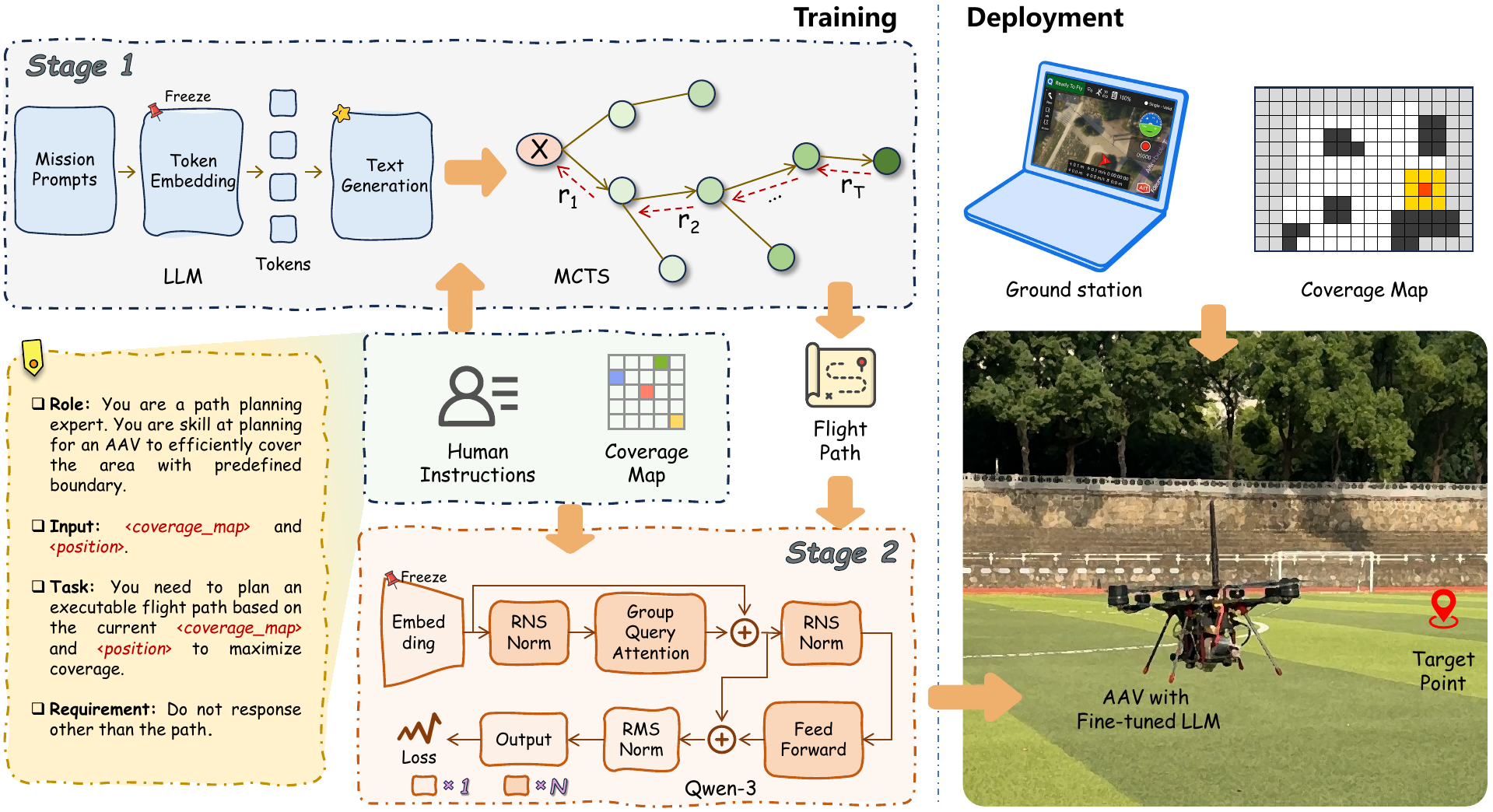}
	
	\caption{
		Overall framework of Skypilot. The training process consists of two stages. In Stage 1, MCTS-based trajectory generation explores the action space and builds high-quality trajectory datasets. In Stage 2, the Qwen3-4B model undergoes full-parameter fine-tuning to enhance inference efficiency. During deployment, the LLM processes the coverage map $\mathcal{M}_t$ and human instructions $\mathcal{I}_t$ to generate the path plan $\mathcal{P}_t$ for coverage missions.
	}
	\label{fig:framework}
\end{figure*}

\section{Problem formulation}
\label{pf}

In our coverage scenario, a single AAV systematically explores a two-dimensional grid environment cluttered with obstacles while responding to natural language instructions from the human operator (Fig.~\ref{fig:scenario}). Equipped with radar-based environmental perception module, the AAV detects obstacles and generates coverage maps, which together with human instructions serve as inputs for decision-making.   The AAV must achieve comprehensive area coverage while avoiding obstacles and adapting to evolving mission priorities in real time.  Our goal is to develop an LLM-enhanced coverage control framework that processes both radar-generated coverage maps and human instructions to generate adaptive path plans while maintaining efficient coverage performance.
We formulate this single-AAV coverage task as a deterministic, continuous-control coverage problem characterized by the tuple $\langle \mathcal{W},\, \mathcal{M}_{t},\, \mathcal{G}^{\star}_t,\, \mathcal{P}_t,\, \mathcal{I}_t\rangle$:

\begin{itemize}
	\item \textbf{Workspace $\mathcal{W}$}: A static, 2D grid environment $\mathcal{W}\subset\mathbb{Z}^2$, partitioned into two disjoint subsets: a traversable free space, $\mathcal{W}_{\text{free}}$, and a non-traversable obstacle space, $\mathcal{W}_{\text{obs}}$, such that $\mathcal{W}=\mathcal{W}_{\text{free}}\cup\mathcal{W}_{\text{obs}}$ and $\mathcal{W}_{\text{free}}\cap\mathcal{W}_{\text{obs}}=\varnothing$.

	\item \textbf{Coverage map $\mathcal{M}_t$}:	A boolean grid map with the same size as  $\mathcal{W}$. It tracks cell visitation status and is dynamically updated in real-time based on environmental information perceived by the AAV. All cells are initialized to \texttt{unvisited} at $t=0$.
	
		\item \textbf{Human instructions $\mathcal{I}_t$}:  Natural language commands received at time 
		$t$ that specify the desired coverage strategy, e.g., \textit{complete coverage}, \textit{rapid traversal}, or \textit{focused area exploration}.

	\item \textbf{Path plan $\mathcal{P}_t$}: Waypoint sequence $\mathcal{P}_t = \langle g_t,\, g_{t+1},\, \dots,\, g_{t+L} \rangle$ generated by the LLM with four-connectivity constraints, where $g_{t+i}$ for $i= 0,1,...,L$ denotes the target trajectory points.
	
	\item \textbf{Coverage objective $\mathcal{G}^{\star}_t$}: The  core mission objective for AAV to determine an optimal path plan $\mathcal{P}_{t}^{\star}$ that aligns with human instructions $\mathcal{I}_t$, expressed as:

	\begin{equation}
	\mathcal{P}_{t}^{\star}=\underset{\mathcal{P}_{t}}{\arg \max }~ \text{Score}(\mathcal{P}_{t}\mid \mathcal{M}_{t}, \mathcal{I}_t),
	\end{equation}
where $\text{Score}(\cdot)$ is the scoring function that evaluates the quality of the path plan \(\mathcal{P}_t\). 
	
\end{itemize}

The system operates through a sequential planning architecture where the AAV iteratively updates its mission execution based on sensory information and human instructions. At time $t$, given the current coverage map $\mathcal{M}_t$, AAV pose $p_t$, and human instructions $\mathcal{I}_t$, the LLM planner solves the coverage objective $\mathcal{G}^{\star}_t$ to generate a  path plan $\mathcal{P}_t$. 
Then, AAV takes the first element $g_t$ from $\mathcal{P}_t$ as the desired waypoint and navigates to that position via its onboard flight control system.
 As the AAV reaches  $g_t$, the coverage map $\mathcal{M}_{t+1}$ is updated to reflect newly visited areas, and the LLM planner generates the next path plan $\mathcal{P}_{t+1}$.
Through this iterative planning-execution-update process, the AAV can effectively adapt to dynamic mission demands, thereby ensuring compliance with the human-specified coverage objectives.

\section{METHODOLOGY}
\label{me}

In this section, we propose our LLM-based AAV coverage search framework, Skypilot. As illustrated in Fig.~\ref{fig:framework}, the overall framework adopts a training-and-deployment paradigm. The training process can be divided into two stages. In the first stage, we employ MCTS to systematically explore promising flight paths and generate high-quality trajectory segments. These segments serve as training data to fine-tune the LLM. In the second stage, we perform full-parameter fine-tuning on the Qwen3-4B model to enhance inference speed while maintaining solution quality. During deployment, the fine-tuned LLM directly generates executable flight path plan $\mathcal{P}_t$ based on coverage map $\mathcal{M}_t$ and human instructions $\mathcal{I}_t$, thereby enabling efficient autonomous exploration.

\subsection{Radar-Based Localization}

To infer the AAV's position in real time, we simulate a mapping-localization module that takes the latest radar beam lengths as input and returns the vehicle's Cartesian position.  
Let $\mathbf{d}=\bigl[d_{1},\,d_{2},\dots,d_{N}\bigr]^{T}$ denote the range vector measured by an $N$-beam 2-D radar mounted at the AAV's centre of mass, where $d_{i}$ is the length of the $i$-th beam in the sensor frame.  Under the assumption of a locally planar environment, each valid beam endpoint lies on the ground plane and is transformed into the world frame via the current attitude estimate.   Then we minimise a geometric residual between these transformed endpoints and a prebuilt grid map:
\begin{equation}
\widehat{ p}
=\arg\min_{ p}
\sum_{i=1}^{N}
\bigl\lVert
g\bigl( R( p)\,[r_i,0]^{ T}+ t( p)\bigr)
\bigr\rVert_2^{2},
\end{equation}
where  
$ p=[x,\,y]^{ T}$ is the horizontal AAV position to be solved,  
$ R( p)$ and $ t( p)$ are the rotation and translation induced by $ p$, and  
$ g(\cdot)$ is the signed-distance field (SDF) of the occupancy grid.

\begin{figure*}[t]
	\centering
	\includegraphics[width=0.99\linewidth]{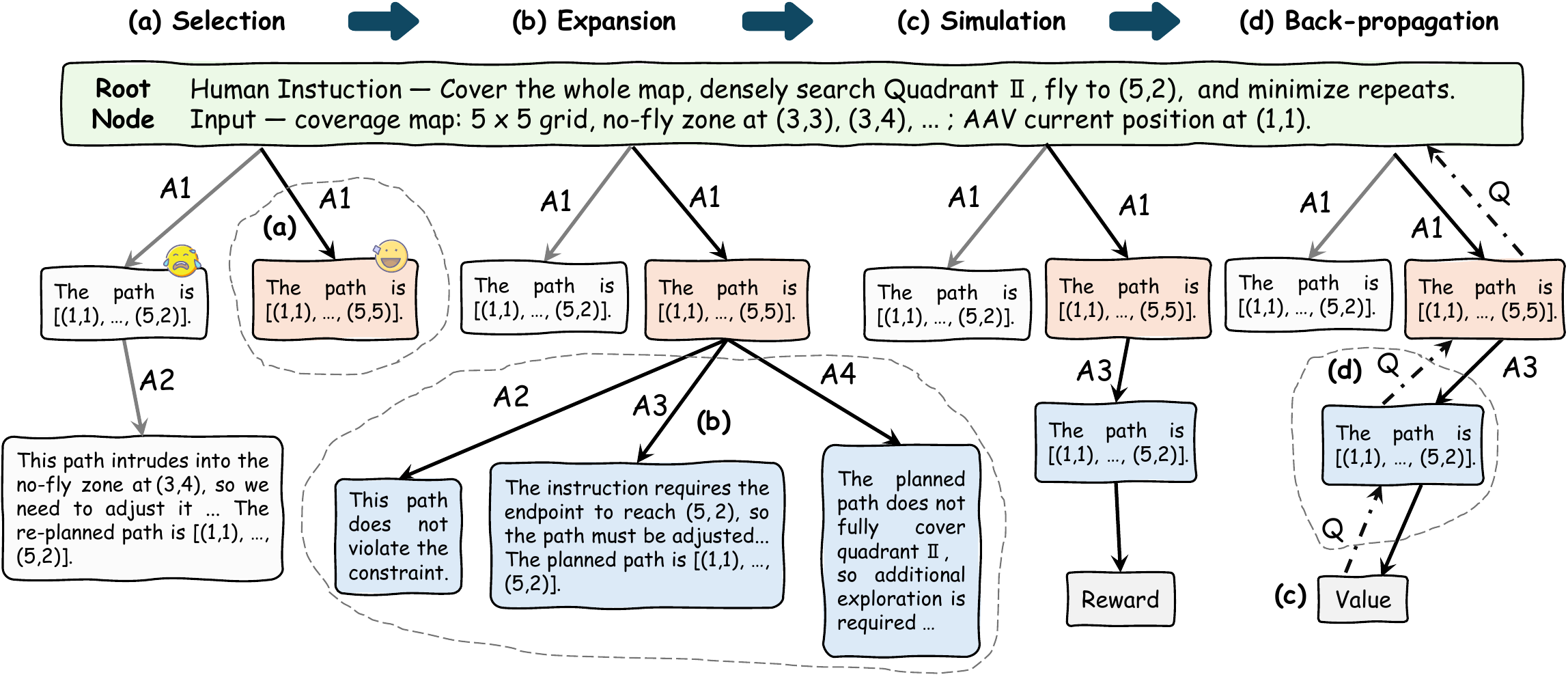}
	
	\caption{Monte Carlo Tree Search process for LLM-based trajectory generation. The four-phase process includes: (a) Selection using UCT to identify promising nodes, (b) Expansion through four action types, (c) Simulation to assess path quality based on coverage ratio and revisit penalty, and (d) Back-propagation to update node values along the search path.}
	\label{fig:MCTS_implement}
\end{figure*}

\subsection{Monte Carlo Tree Search}
\label{sec:mcts}
LLMs face challenges in directly supporting decision-making tasks due to the lack of real-world experience, especially in tasks with specific physical constraints \cite{ahn2022can}. Hence, we employ MCTS \cite{zhang2024accessing} to dynamically balance exploration and exploitation, thereby enabling LLMs to generate feasible and high-quality coverage paths.

Formally, for a given input $x=(\mathcal{M},\mathcal{I}_t)$, MCTS gradually constructs a search tree $T_r$, where each node corresponds to a candidate trajectory segment generated by the LLM under specific actions. A path from the root node to a leaf node $s_i$ forms a potential solution trajectory $\tau = s_0 \oplus s_1 \oplus \dots \oplus s_d$, where $s_d$ is the terminal node. After multiple rollouts, a set of candidate trajectories $\mathcal{T}=\{\tau_1, \tau_2,\dots,\tau_n\}$ can be extracted, from which we identify the optimal trajectory. As shown in Fig.~\ref{fig:MCTS_implement},  a rollout in MCTS consists of \textit{selection}, \textit{expansion}, \textit{simulation}, and \textit{back-propagation}.

\subsubsection{Selection}
In the selection phase of MCTS, the algorithm traverses the tree from the root to select a promising node for expansion. To balance between exploration and exploitation, we employ the upper confidence bounds applied to trees to select each child node~\cite{kocsis2006bandit}:
\begin{equation}
\label{eq:UCT}
UCT=Q(s)+\omega\sqrt{\frac{\ln N(\mathrm{parent}(s))+1}{N(s)+\epsilon}},
\end{equation}
where $Q(s)$ represents the estimated value of the node $s$; $N(s)$ and $N(\mathrm{parent}(s))$ denote the visit count of node $s$ and its parent; $\omega$ indicates the exploration parameter that adjusts the weight between exploration and exploitation.

\subsubsection{Expansion}

In MCTS, the expansion phase broadens the search tree by adding new action branches. However, most MCTS-based methods for LLMs rely on a single type of action, which limits exploration and can trap the LLM in the search deadlock~\cite{qi2025mutual}. In the context of LLM-based flight planning, an additional challenge is that the generated trajectories often violate operational constraints, stemming from the model's lack of domain-specific knowledge. To address these issues, we introduce four distinct types of actions to diversify the search space and improve the planned trajectory feasibility:

\begin{itemize}
	\item \textbf{A1:} Generate trajectory. This action acts on the root node and prompts the LLM to generate the coverage path based on the initial input $x$.
	
	\item\textbf{A2:} Regenerate trajectory. When a trajectory violates a hard constraint, this action is triggered. The LLM regenerates the entire coverage path for the current node $s_i$ based on specific, automated feedback $f$ (e.g., ``Error: path enters no-fly zone at coordinates $(m, n)$'').
	
	\item \textbf{A3:} Finetune trajectory. This action refines the path at node $s_i$ to better meet a qualitative objective from the human instruction $\mathcal{I}$ (e.g., ``Pass through this area quickly'' or ``Search the quadrant II carefully''). It focuses on local, goal-oriented adjustments rather than wholesale correction.
	
	\item \textbf{A4:} Evaluation trajectory. For any given node $s$, this action assesses compliance with flight objectives and constraints, then determine if further exploration is required.
\end{itemize}

The four actions $\{\textrm{A1},\textrm{A2},\textrm{A3},\textrm{A4}\}$ define the action space $\mathcal{A}$ for the search tree. During each iteration, the MCTS selects a leaf node using UCT and select an action $a_i$ from $\mathcal{A}$. The chosen action $a_i$ then serves as the prompt for the LLM to generate the subsequent node $s_{i+1}$ based on the current node $s_i$ and the input $x$.

\subsubsection{Simulation}
The simulation stage of MCTS is responsible for estimating the Q-value of the state-action pair $(s_{i+1}, a_{i})$. While previous MCTS-based methods have employed LLMs to directly assign scalar value~\cite{hao2023reasoning,zhang2024accessing}, the models' limited grasp of the physical world often leads to inaccurate quality assessments. To address this, we design a physical-informed reward function that combines objective coverage metrics with LLM-evaluated instruction compliance. The function is defined as follows:
\begin{equation}
\label{eq:updateQ}
Q(s) = 
\begin{cases} 
c_1  \frac{\left|\mathcal{C}_\text{visited}(s)\right|}{\left|\mathcal{C}_\text{free}\right|} 
\!- \!c_2 \frac{\left|\mathcal{C}_\text{revisited}(s)\right|}{\left|\mathcal{C}_\text{visited}(s)\right|} 
\!+\! c_3 \mathbb{E}(s, I_t), 
& s \in \mathcal{S}_\text{valid} \\[6pt]
0, & \text{otherwise}
\end{cases},
\end{equation}
where $c_1, c_2, c_3 > 0$ are weighting coefficients; $s$ is a node in the search tree corresponding to a generated flight trajectory; $\mathcal{S}_\text{valid}$ denotes the set of all trajectories satisfying mandatory hard constraints; the sets $\mathcal{C}_\text{free}$, $\mathcal{C}_\text{visited}(s)$, and $\mathcal{C}_\text{revisited}(s)$ represent the free, visited, and revisited cells respectively; 	and $\mathbb{E}(s, I) \in [0, 1]$ quantifies the instruction compliance score assigned by the LLM based on the semantic alignment between trajectory $s$ and human directives $I_t$. This formulation balances quantitative coverage efficiency (coverage rate minus revisit penalty) with qualitative instruction adherence assessed by the language model.

\subsubsection{Back-Propagation}
Upon the evaluation of the newly expanded leaf node, a back-propagation process is to update the value of all nodes along the trajectory from the new node to the root. This process ensures that the value of a node not only reflects its own objective quality but also incorporates the optimal potential of future trajectories extending from it. The update for each ancestor node $s$ is performed as follows:
\begin{equation}
\label{eq:back}
\widehat{Q}(s) = (1 - \alpha)Q(s) + \alpha \max\nolimits_{s' \in \mathrm{Children}(s)} Q(s'),
\end{equation}
where $\widehat{Q}(s)$ denoted the updated value estimate and $Q(s)$ is the current value estimated of the node $s$; $\max\nolimits_{s' \in \mathrm{Children}(s)} Q(s')$ represents the highest Q-value among the children of the node being updated; $\alpha \in (0, 1]$ indicates the learning parameter.

\subsection{Full Parameter Fine-tuning}

\begin{algorithm}[h]
	\caption{	Training  of Skypilot}
	\label{alg:skypilot}
	\textbf{Input:}  Empty Dataset $ \mathcal{D} $,  GPT-4o $\mathcal{\phi}_{\textrm{gpt}}$, original Qwen3-4B  $\mathcal{\phi}_{\textrm{llm}}$,    set of candidate trajectories $\mathcal{T}$, action space $\mathcal{A}$ with associated  prompts $(p_{\textrm{generate}},\, p_{\textrm{regenerate}},\, p_{\textrm{finetune}},\, p_{\textrm{evaluate}})$,  maximum data collection episodes $\mathcal{N}_{e}$, maximum  number of rollouts $\mathcal{N}_r$\\
	\textbf{Output:} Fine-tuned parameters $\theta^{\star}$
	\begin{algorithmic}[1]
		\Statex 	\textbf{Stage 1: Data collection}
		\For{$i\leftarrow1$ to  $\mathcal{N}_{e}$}
		\State Initialize  coverage tuple $\langle \mathcal{W},\, \mathcal{M},\, \mathcal{G}^{\star},\, \mathcal{I}\rangle$;
		\State Initialize  search tree $T_r$ with root node $s_0 = \phi_{\textrm{gpt}}(\mathcal{P} \mid\mathcal{M}, \mathcal{I}, p_{\textrm{generate}})$;
		\State  $\mathcal{T} \leftarrow \{\}$;
		\For{$j\leftarrow1$ to $\mathcal{N}_r$ }
		\State $\tau_j \leftarrow \{s_0\}$, $s \leftarrow s_0$;
		\While{$s$ is not a terminal state}
		\State $N(s) \leftarrow N(s) + 1$;
		\State Randomly select an action $a$ from $\mathcal{A}$;
		\If{$a  =  \text{A2}$} \Comment{Regenerate with feedback}
		\State $s'=\phi_{\textrm{gpt}}(\mathcal{P}_{t}'\mid\mathcal{M}_t, \mathcal{I}_t,p_{\textrm{regenerate}},\mathcal{P}_t,f)$;
		\ElsIf{$a = \text{A3}$} \Comment{Fine-tune trajectory}  
		\State $s'=\phi_{\textrm{gpt}}(\mathcal{P}_{t}'\mid\mathcal{M}_t, \mathcal{I}_t,p_{\textrm{finetune}},\mathcal{P}_t)$;
		\ElsIf{$a = \text{A4}$} \Comment{Evaluate trajectory}
		\State $s'=\phi_{\textrm{gpt}}(\mathcal{P}_{t}'\mid\mathcal{M}_t, \mathcal{I}_t,p_{\textrm{evaluate}},\mathcal{P}_t,f)$;
		\EndIf
		\State Evaluate  $s'$ and compute its value by Eq.~(\ref{eq:updateQ}); 
		\State Propagate value updates to ancestor nodes of \textcolor{white}{\qquad}\textcolor{white}{\hspace{4.1em}}   $s'$ by Eq.~(\ref{eq:back});
		\State $\tau_j \leftarrow \tau_j \cup \{s'\}$;
		\State Select the next node $ s'' $ to explore by Eq.~(\ref{eq:UCT});
		\State  $s \leftarrow s''$;
		\EndWhile
		\State $\mathcal{T} \leftarrow \mathcal{T} \cup \{\tau_j\}$;
		\EndFor
		\State Extract optimal trajectory $\tau^{*}$ from $\mathcal{T}$; 
		\State Add $(\mathcal{M}, \mathcal{I}, \tau^{*})$ to $\mathcal{D}$;
		\EndFor
		\Statex  \textbf{Stage 2: Full-parameter fine-tuning}
		\makeatletter
		\setcounter{ALG@line}{0}%
		\makeatother
		\State Tokenize each training sample  $(\mathcal{M}, \mathcal{I}, \tau^{*})$  in $ \mathcal{D}$;
		\State Construct tokenized dataset $\mathcal{D'}$ by Eq.~(\ref{eq:D});
		\For{$t\leftarrow1$ to $T$}
		\State Sample mini-batch from  $\mathcal{D'}$;
		\State Calculate loss $ \mathcal{L}(\theta_{t-1})  $ by Eq.~(\ref{eq:loss});
		\State Compute gradient $g_t=\nabla_{\theta_{t-1}}\mathcal{L}(\theta_{t-1})$;
		\State Update moments $m_t,v_t$ by Eq.~(\ref{eq:updatemoment});
		\State Update Qwen3-4B parameters $\theta_t $  by Eq.~(\ref{eq:update_theta});
		\EndFor
		\State $\theta^{\star}\leftarrow\theta_{T}$;
	\end{algorithmic}
\end{algorithm}

Current approaches to generating flight paths for AAVs that leverage LLMs with MCTS for inference are computationally expensive. This computational burden  arises because the MCTS framework inherently requires numerous iterative rollouts to explore the action space and converge on optimal solutions. To substantially reduce inference time for practical AAV operations, we perform full-parameter fine-tuning on Qwen3-4B.  
The  fine-tuning process begins with  constructing a high-quality dataset $\mathcal{D'}$ of 23,000 trajectory-planning samples:
\begin{equation}
\label{eq:D}
\mathcal{D'} = \bigcup_{k=1}^{K} \left\{ \left( x_i^{k}, y_i^{k} \right) \right\}_{i=1}^{n_k},
\end{equation}
where $K$ is the number of dataset batches; $n_k$ is the number of samples in the $k$-th subset; 
$x_{i}^{k}=(\mathcal{M}_k,\mathcal{I}_k)_{i}$ represents the $i$-th input token in the $k$-th sequence; 
$y_{i}^{k}$ denotes the output token, generated using GPT-4o in conjunction with MCTS, 
as detailed in Section~\ref{sec:mcts}. 
Then,  Qwen3-4B is  trained by minimizing the  negative log-likelihood loss $\mathcal{L}(\theta)$, which can be formally defined as~\cite{xiao2025fm}
\begin{equation}
\label{eq:loss}
	\mathcal{L}(\theta) = - \frac{1}{K} \sum_{k=1}^{K} \frac{1}{n_k} \sum_{i=1}^{n_k} 
	\log P(y_{i}^{k} \mid y_{\leq i}^{k}, x_{i}^{k}; \theta),
\end{equation}
where $\theta$ is the model parameter. To enhance the model's generalization capability,  we employ AdamW optimizer~\cite{loshchilov2017decoupled}  for parameter updates:
\begin{equation}
\label{eq:update_theta}
\theta_t = \theta_{t-1} - \eta \left( \frac{\hat{m}_t}{\sqrt{\hat{v}_t} + \epsilon} + \lambda \theta_{t-1} \right), 
\end{equation}
where $\eta$ denotes the learning rate, which is dynamically adjusted by the learning rate scheduler; $\lambda$ represents the weight decay rate; $\hat{m}_t=m_t/(1 - \beta_1^t)$ and $\hat{v}_t=v_t/(1 - \beta_2^t)$ are the bias-corrected estimates of the first and second moments, respectively. The raw moments $\hat{m}_t$ and $\hat{v}_t$ are updated via
\begin{equation}
\label{eq:updatemoment}
\begin{gathered}
m_t = \beta_1 m_{t-1} + (1 - \beta_1) g_t \\
v_t = \beta_2 v_{t-1} + (1 - \beta_2) g_t^2
\end{gathered},
\end{equation}
where $g_t$ represents the accumulated gradient; $\beta_1$ and $\beta_2$ denote the momentum coefficients. To optimize memory utilization, we adopt ZeRO-3 mechanism~\cite{rajbhandari2020zero} to shard the optimizer states $\Psi$, gradients $G$, and model parameters $\theta$ across devices:
\begin{equation}
X = [X^{(1)}, X^{(2)}, \dots, X^{(N_d)}], \quad \forall X \in \{\Psi, G, \theta \}, 
\end{equation}
where $N_d$ is the number of training devices.

\subsection{Summary and Analysis}
The complete training procedure  of the proposed Skypilot approach is summarized in Algorithm~\ref{alg:skypilot}. It consists of two main stages. The data collection stage leverages GPT-4o to generate high-quality trajectory samples through systematic tree search exploration, where different actions (regeneration, fine-tuning, evaluation) are applied to refine path planning decisions (Line 7-22 of Stage 1). Each episode produces an optimal trajectory that captures the mapping between coverage states, human instructions, and effective path plans. The collected dataset is then used to fine-tune the lightweight Qwen3-4B model (Line 3-9 of Stage 2), enabling it to perform autonomous coverage planning without requiring the computational overhead of GPT-4o during deployment.

Our Skypilot framework delivers key practical benefits for AAV deployment. The offline trajectory generation approach enables efficient operation on resource-constrained platforms, while MCTS exploration during training produces robust policies through comprehensive state-action coverage. Although training complexity depends on episode count and rollout depth, the deployed model achieves consistent inference performance across different environments. This design supports scalable mission deployment while maintaining reliable real-time execution.

\section{EXPERIMENTS}
\label{exp}
In this section, we evaluate Skypilot through simulation benchmarks and physical experiments in indoor and outdoor environments.

\begin{table*}[hbtp]
	\setlength{\tabcolsep}{3.3pt}
	\renewcommand{\arraystretch}{1.5}
	\centering
	\caption{Performance of eight LLM-Based Planners on Single AAV Coverage Across Varying Obstacle Densities}
	\label{tab:llm_performance}
	\begin{tabular}{cccc ccc ccc}  
		\toprule
		\multirow{2}{*}[-0.8ex]{Model} &
		\multicolumn{3}{c}{Sparse Obstacles} &
		\multicolumn{3}{c}{Medium Obstacles} &
		\multicolumn{3}{c}{Dense Obstacles} \\
		\cmidrule(lr){2-4}\cmidrule(lr){5-7}\cmidrule(lr){8-10}
		& CR (\%) $\uparrow$  & DR (\%) $\downarrow$& CSI (\%) $\uparrow$ 
		& CR (\%) $\uparrow$ & DR (\%) $\downarrow$& CSI (\%) $\uparrow$
		& CR (\%) $\uparrow$& DR (\%) $\downarrow$& CSI (\%) $\uparrow$\\
		\midrule
		Gemma-7B-it~\cite{gemmateam2024gemmaopenmodelsbased}                 & 60.84$\pm$26.54 &   4.77$\pm$11.36 &  51.10$\pm$22.30 & 55.11$\pm$26.05 &   6.63$\pm$13.83 &  39.68$\pm$18.76 & 54.23$\pm$20.21 &   3.47$\pm$8.68 &  29.28$\pm$10.91\\
		Phi-3-mini-4B~\cite{abdin2024phi3technicalreporthighly}               & 71.68$\pm$22.74& 7.16$\pm$11.03& 35.84 $\pm$11.37  & 62.82$\pm$18.89& 6.07$\pm$10.37 & 25.13$\pm$7.56 & 61.81$\pm$25.44 &   7.97$\pm$10.96 &  28.43$\pm$11.70 \\
		Mistral-7B-Instr.~\cite{jiang2023mistral7b}           &  64.33$\pm$21.56 &  5.90$\pm$13.75 & 37.31$\pm$12.51  & 61.13$\pm$21.88 &   6.17$\pm$15.12 &  24.45$\pm$8.75& 56.81$\pm$22.68 &   4.14$\pm$7.26 &  23.86$\pm$9.53 \\
		Llama-3-8B-Instr.~\cite{meta2024llama3}           & 75.88$\pm$19.47 &  8.07$\pm$14.04 & 31.87$\pm$8.18  & 71.75$\pm$17.07 &   9.55$\pm$12.85 &  38.75$\pm$9.22 & 69.68$\pm$18.30& 8.52 $\pm$16.32 & 27.87$\pm$7.32 \\
		Qwen3-Emb-8B~\cite{yang2025qwen3technicalreport}                & 82.06$\pm$21.81 & 1.85$\pm$4.83 &  57.44 $\pm$15.27  & 79.02$\pm$21.36 & \textbf{1.72$\pm$4.98} & 61.63$\pm$16.66 & 75.83$\pm$24.88 &   1.69$\pm$5.57 &  53.08$\pm$17.42 \\
		Qwen3-Emb-4B~\cite{yang2025qwen3technicalreport}                & 74.19$\pm$26.37 &  4.40$\pm$10.51 & 46.00$\pm$16.35 &73.72$\pm$24.73 &   4.13$\pm$8.81 &  54.55$\pm$18.30& 71.30$\pm$24.73 & 2.17$\pm$6.30 & 38.50$\pm$13.35 \\
		GPT-4o~\cite{openai2022chatgpt}                       & 84.44$\pm$17.82&  4.38$\pm$12.72& 50.66$\pm$10.69 & 84.07$\pm$17.48& 1.77$\pm$6.39 & 35.31$\pm$7.34 & 79.50$\pm$14.71& 6.04$\pm$13.05 & 20.67$\pm$3.82 \\
		\textbf{Skypilot (Ours)}    & \textbf{91.98$\pm$13.51} & \textbf{1.35$\pm$6.05} & \textbf{84.62$\pm$12.43} & \textbf{88.99$\pm$14.36} & 1.73$\pm$8.01 & \textbf{71.19$\pm$11.49} & \textbf{84.76$\pm$19.70} & \textbf{1.50$\pm$5.88} & \textbf{59.33$\pm$13.79} \\
		\bottomrule
	\end{tabular}
	\label{llm_performance}
\end{table*}

\subsection{Simulation Setup}
\subsubsection{Parameter Settings}
To assess the performance of Skypilot, we conducted comprehensive simulation experiments. 
Our experimental framework comprises two sequential stages: first, we collect training data through MCTS simulations powered by GPT-4o, then perform supervised fine-tuning on the Qwen3-4B model.
In the data collection phase, we generated trajectory samples across diverse coverage scenarios, accumulating over 23,000 episodes with multiple random seeds to ensure statistical robustness and scenario diversity. For the fine-tuning stage, we employed full-parameter optimization using the AdamW optimizer with a learning rate of 1.0$\times$10$^{-5}$ and a cosine annealing scheduler incorporating a 10\% warmup ratio. To optimize computational efficiency, we utilized ZeRO Stage 3~\cite{rajbhandari2020zero} optimization for memory management and BFloat16~\cite{bfloat162019} mixed precision training. The effective batch size was set to 2 through gradient accumulation. The complete training procedure required approximately three hours when executed on a distributed system equipped with 4 NVIDIA RTX 4090 GPUs.

\subsubsection{Performance Metrics}	
To comprehensively evaluate coverage planning performance, here we define four metrics:

\begin{itemize}
	\item \textbf{Coverage rate (CR)}: The percentage of free cells visited by the AAV during a mission, defined as:
	\begin{equation}
	\text{CR} = \frac{|\mathcal{C}_{\text{visited}}|}{|\mathcal{C}_{\text{free}}|} \times 100\%,
	\end{equation}
	where $\mathcal{C}_{\text{visited}}$ denotes the set of visited cells and $\mathcal{C}_{\text{free}}$ represents all traversable cells in the map.
	
	\item \textbf{Duplicate rate (DR)}: The percentage of redundant cell visits, measuring path efficiency:
	\begin{equation}
	\text{DR} = \frac{|\mathcal{C}_{\text{revisited}}|}{|\mathcal{C}_{\text{visited}}|} \times 100\%,
	\end{equation}
	where $\mathcal{C}_{\text{revisited}}$ contains cells visited more than once.
	
	\item \textbf{Coverage-success index (CSI)}: A composite metric that penalizes failed missions due to collision:
	\begin{equation}
	\text{CSI} = \text{CR} \times \text{SR},
	\end{equation}
	where SR (success rate) represents the fraction of trials completed without obstacle collisions. A mission is marked as failed if the planned trajectory intersects with any obstacle.
	
	\item \textbf{Inference latency (IL)}: The average computational time (in seconds) required to generate a complete coverage trajectory, measured from input reception to path output.
\end{itemize}

All metrics are computed as averages over 50 independent test runs to ensure statistical reliability.

\subsection{Comparative Results}

We conducted comparative experiments against seven state-of-the-art LLM-based planning methods across three distinct obstacle density configurations (sparse, medium, and dense). Each configuration was tested over 50 episodes to ensure statistical reliabilitys.  The baseline methods include 	Gemma-7B-it~\cite{gemmateam2024gemmaopenmodelsbased}, 		Phi-3-mini-4B~\cite{abdin2024phi3technicalreporthighly},  		Mistral-7B-Instr~\cite{jiang2023mistral7b},  Llama-3-8B-Instr~\cite{meta2024llama3}, GPT-4o~\cite{openai2022chatgpt},  and Qwen3 variants (Qwen3-Emb-4B, Qwen3-Emb-8B)~\cite{yang2025qwen3technicalreport}, representing diverse architectural approaches and model scales in current LLM-based planning research.

Table~\ref{llm_performance} presents the comprehensive performance evaluation results. Among all evaluated models, Skypilot achieves the highest overall performance across all environments. In sparse obstacles, it attains 91.98\% coverage rate with minimal duplication (1.35\%) and superior CSI (84.62\%). As obstacle density increases,  Skypilot maintains its competitive advantage, achieving 88.99\% (medium) and 84.76\% (dense) coverage rates while sustaining superior CSI values, demonstrating robust adaptability to environmental complexity.

Following Skypilot, GPT-4o demonstrates remarkable consistency with coverage rates of 84.44\%, 84.07\%, and 79.50\% across the three configurations, though this stability comes with higher duplicate rates ranging from 4.38\% to 6.04\%. In contrast, models such as Gemma-7B-it and Llama-3-8B-Instr show significant performance degradation in dense environments, with CSI values dropping to 29.28\% and 27.87\% respectively. This reveals critical limitations in collision avoidance capabilities under challenging conditions. Similarly, Phi-3-mini-4B and Mistral-7B-Instr struggle with complex environments, achieving CSI values of only 28.43\% and 23.86\% in dense configurations.

The Qwen3-Emb variants demonstrate notably efficient path planning capabilities, with Qwen3-Emb-8B maintaining exceptionally low duplicate rates  across all scenarios while achieving reasonable coverage rates. Qwen3-Emb-14B shows similar trends with slightly higher duplication but maintains competitive performance, indicating effective trajectory optimization strategies within this model family.

Overall, these results validate that effective LLM-based coverage planning requires balanced optimization of coverage maximization and collision avoidance. Skypilot's consistent superiority across all metrics and environmental configurations demonstrates its effectiveness for autonomous coverage  in increasingly complex operational scenarios.

\begin{figure}[hbtp]
	\centering	\includegraphics[width=0.99\linewidth]{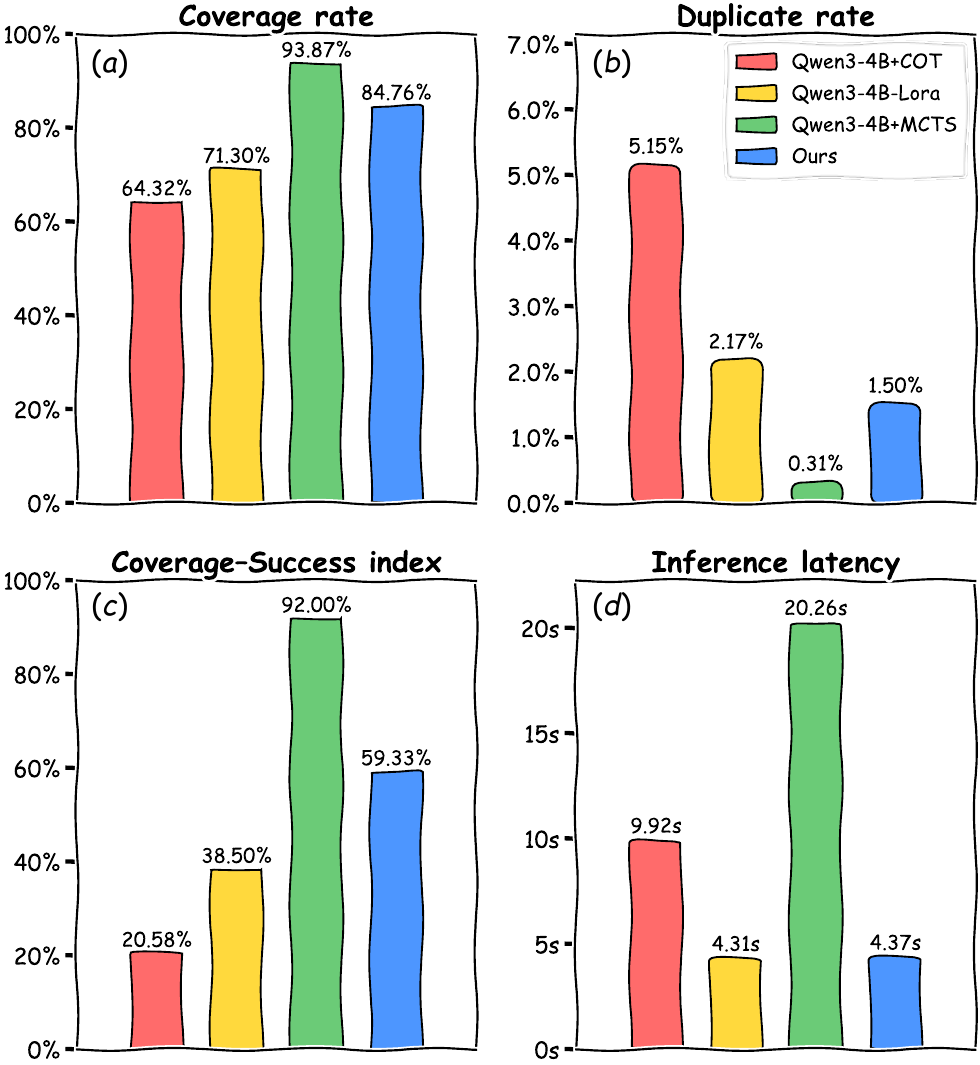}
	
	\caption{Ablation study results of different LLM-based planners in dense obstacle environments. Metrics include coverage rate (CR), duplicate  rate (DR),  coverage success index (CSI), and inference latency (IL). }
	\label{fig:ablation}
\end{figure}

\subsection{Ablation Studies}

To validate the effectiveness of each component in Skypilot, we conducted ablation experiments with four configurations: 
\begin{itemize}
	\item \textbf{Qwen3-4B+CoT}:  original Qwen3-4B model prompted with Chain-of-Thought (CoT) reasoning~\cite{wei2022chain}.
	
	\item \textbf{Qwen3-4B-LoRA}: Qwen3-4B fine-tuned via LoRA~\cite{hu2022lora}.
	
	\item \textbf{Qwen3-4B+MCTS}: LoRA-tuned model augmented with MCTS.%
	
	\item \textbf{Skypilot}: our  proposed framework.%
\end{itemize}

Fig.~\ref{fig:ablation} presents the ablation results, showing the progressive contribution of each component. The baseline Qwen3-4B+CoT achieves 64.32\% coverage rate, which improves to 71.30\% with LoRA fine-tuning, demonstrating a 10.9\% improvement and validating the benefits of task-specific adaptation. Adding MCTS yields dramatic performance gains, achieving 93.82\% coverage rate, 0.31\% duplicate rate, and exceptional 92.00\% CSI, effectively mitigating hallucination issues prevalent in LLM-based planners.

However, MCTS's superior performance incurs substantial computational cost, requiring 20.26s inference time due to multi-round rollouts, making it impractical for real-time  decision-making. In contrast, our complete Skypilot framework achieves 84.76\% coverage rate, 1.50\% duplicate rate, and 59.33\% CSI with only 4.37s inference time. Compared to the baseline, this represents a 31.5\% increase in coverage, 70.9\% reduction in duplicate rate, and 188.4\% improvement in CSI, while achieving 78.4\% reduction in computation time compared to MCTS.

These results validate the meaningful contribution of each component, with Skypilot achieving an optimal balance between performance and computational efficiency for practical deployment.

\subsection{Indoor Experiment Results}

\begin{figure*}[hbtp]
	\centering
	\includegraphics[width=0.99\linewidth]{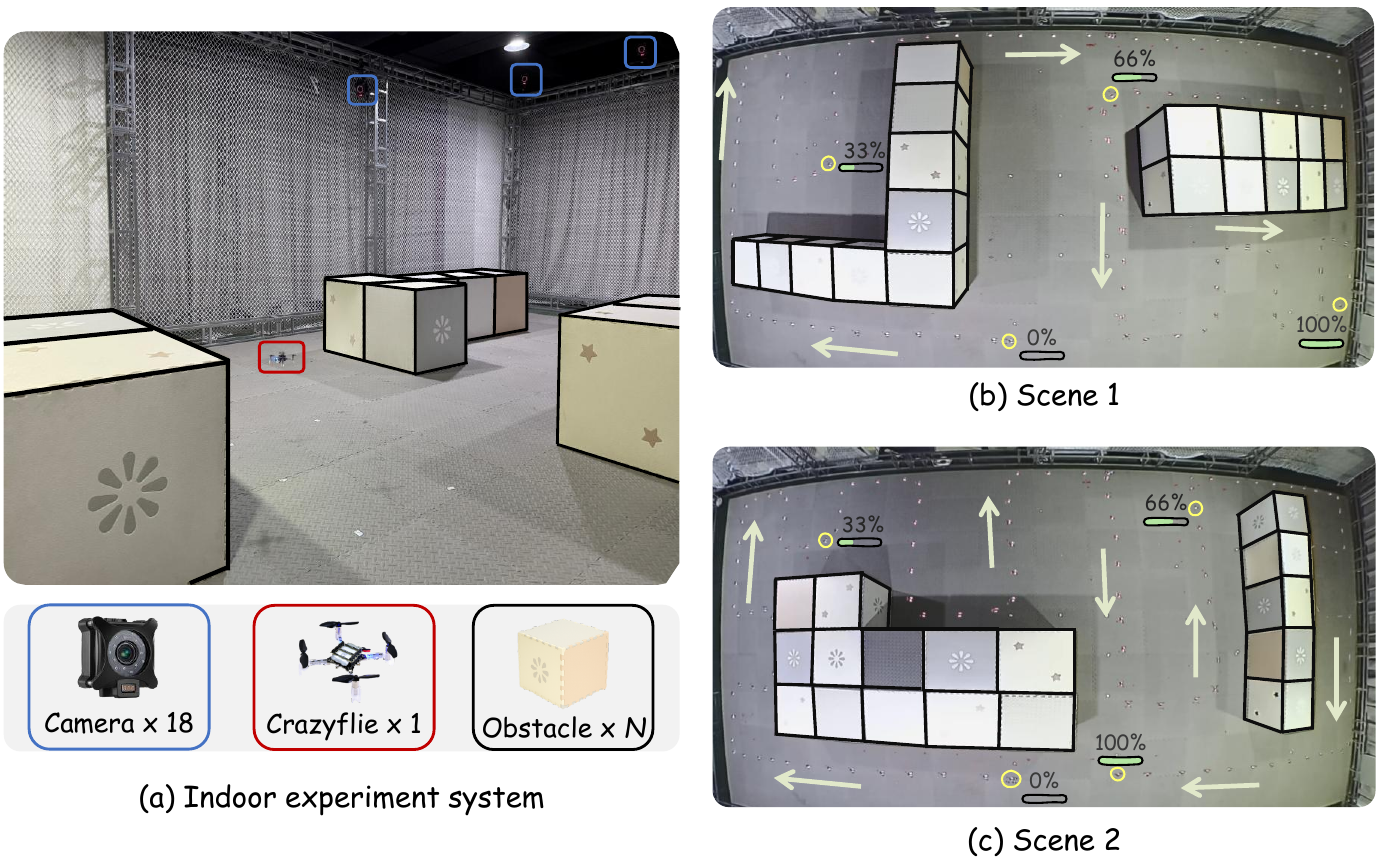}
	
	\caption{Indoor coverage search experiments. (a) Test arena with motion capture system, Crazyflie AAV, and reconfigurable obstacles. (b--c) Two obstacle configurations showing progressive coverage from 0\% to 100\% with overlaid flight trajectories (light-colored arrows), demonstrating the LLM planner's ability to adapt paths in real-time while maintaining complete coverage and obstacle avoidance. The flight video is available at  \url{https://youtu.be/_SEGjsGqFrU}. }
	\label{fig:indoor}
\end{figure*}

\subsubsection{Physical Experimental Setup} As illustrated in Fig.~\ref{fig:indoor} (a), indoor experiments are conducted in an 11 m $\times$ 6.4 m motion-capture arena equipped with 18 cameras. A Crazyflie 2.1 quadrotor navigates among a set of foam-board obstacles that are repositioned between trials to create different obstacle configurations. The camera network provides millimetre-accurate poses to an offboard computer running Skypilot, which transmits position setpoints via 2.4 GHz Crazyradio.

\subsubsection{Time-Lapse Trajectory and Analysis} Fig.~\ref{fig:indoor} (b)--(c) demonstrate the LLM planner's adaptive coverage capabilities across two  obstacle configurations. The workspace contains repositionable modular obstacles, with coverage progress shown at 0\%, 33\%, 66\%, and 100\% completion stages. 
Light yellow arrows  denote flight directions.

In Scene 1 (Fig.~\ref{fig:indoor}(b)), an L-shaped obstacle on the left and a rectangular block on the right create narrow corridors. The Crazyflie starts from the lower-left corner and executes a counter-clockwise coverage pattern, first following the L-shaped wall's contour while maintaining safe clearance. As coverage reaches 33\%, it smoothly transitions through the central corridor, demonstrating precise navigation in confined spaces. The Crazyflie then circumnavigates the right obstacle with controlled movements, gradually expanding its coverage area. Throughout the mission, Skypilot adopts a modified lawnmower pattern to the obstacle geometry, efficiently navigating recesses and tight spaces to achieve complete coverage while minimizing redundancy.

Scene 2 (Fig.~\ref{fig:indoor}(c)) presents a different spatial challenge after obstacle reconfiguration. The blocks form an L-shaped structure on the left boundary and a vertical tower on the right side, creating an asymmetric environment. Despite this layout change, Skypilot automatically adapts without manual replanning or parameter adjustment. Starting from the bottom, the trajectory begins with upward sweeps along the L-shaped facade, systematically covering the central area before addressing boundary regions. As coverage progresses, the planner generates S-turn maneuvers to navigate the L-formation's interior recesses. After 66\% coverage, the Crazyflie executes a spiral pattern around the right tower, ensuring complete coverage near this narrow obstacle. This adaptive behavior demonstrates the LLM's ability to generate context-appropriate strategies without scenario-specific programming.

The comparative analysis reveals Skypilot's key advantage: maintaining 100\% coverage in both configurations while dynamically adjusting to obstacle placement changes. Unlike traditional methods requiring recalibration for unexpected layouts, Skypilot seamlessly transitions between coverage patterns. The system adapts from contour-following in Scene 1 to sweep-and-spiral in Scene 2, demonstrating autonomous adaptive navigation.

\subsection{Outdoor Experiment Results}
\subsubsection{Outdoor Experimental Platform}
\begin{figure}[!h]
	\centering	\includegraphics[width=0.99\linewidth]{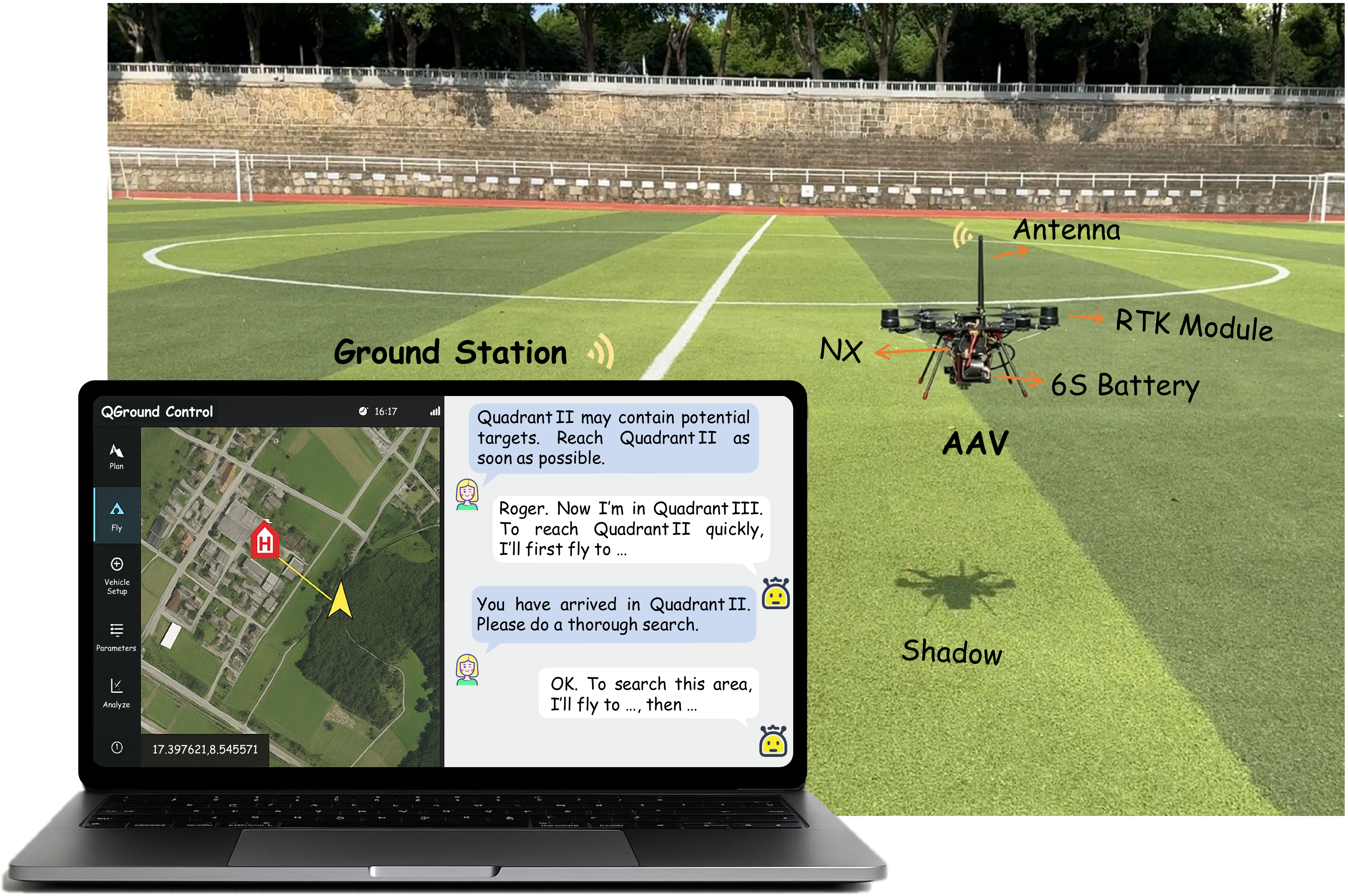}
	\vspace{-10pt} 
	\caption{Outdoor experimental system for AAV coverage search. }
	\label{fig:outdoor_settings}
\end{figure}

\begin{figure*}[!b]
	\centering
	\includegraphics[width=0.99\linewidth]{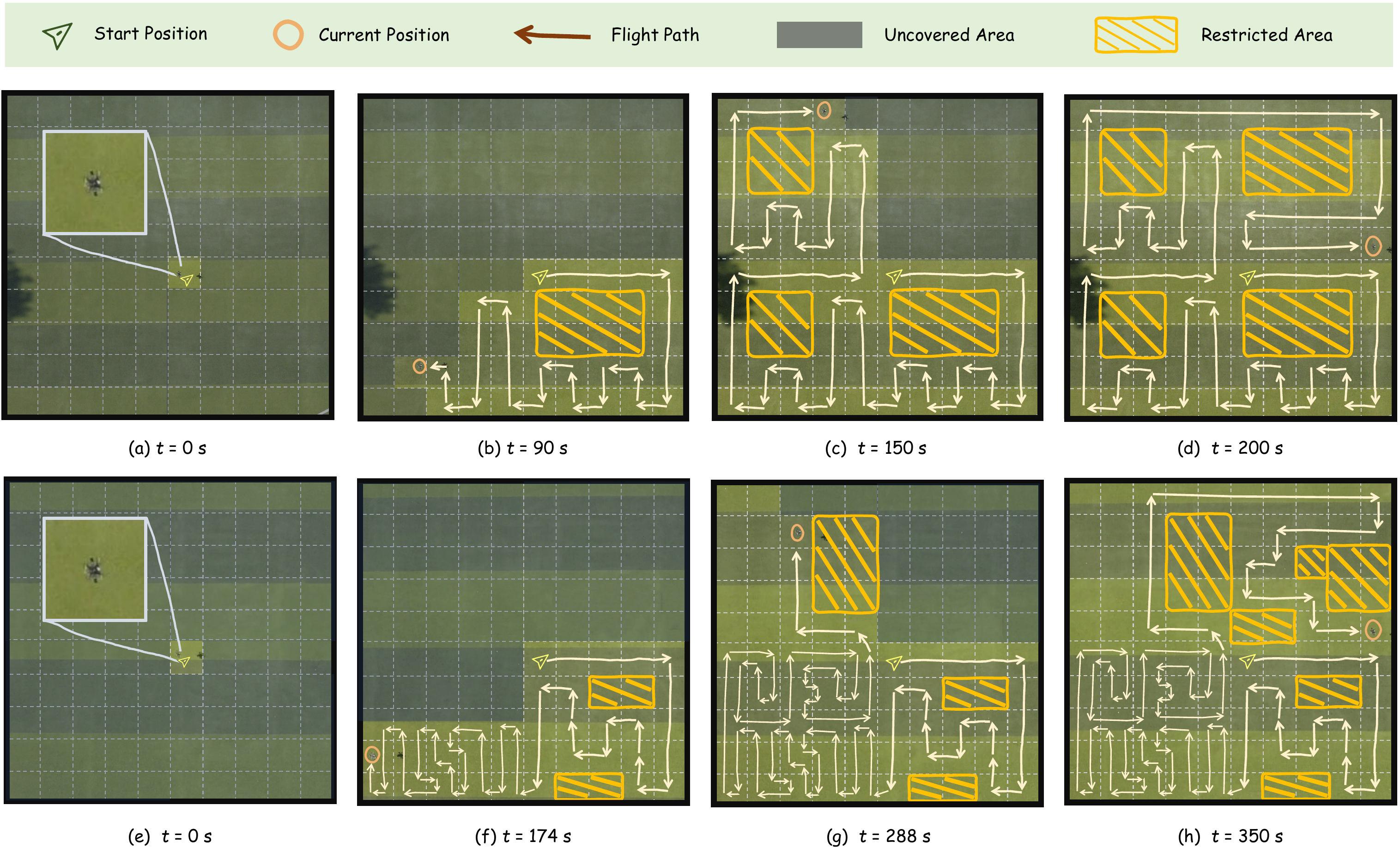}
	\vspace{-4pt} 
	\caption{Key frames of two outdoor single-AAV coverage missions generated by the proposed LLM-based planner. (a)--(d): snapshots at $t = $ 0, 90, 150, and 200 s  show a baseline coverage sweep that incrementally visits all free cells while avoiding  restricted areas.  
		(e)--(h): snapshots at $t$ = 0, 174, 288, and 350 s illustrate human-in-the-loop interaction with the LLM planner, which assigns quadrant-specific subtasks: normal-obstacle-density coverage, fine-grained search, rapid transit, and dense-obstacle coverage. 	The flight video is available at  \url{https://youtu.be/_SEGjsGqFrU}.	}
	\label{fig:outdoor}
\end{figure*}

Fig.~\ref{fig:outdoor_settings} shows the outdoor experimental platform. The AAV features four 7-inch tri-blade propellers, a 6S 10,000 mAh LiPo battery, RTK-GNSS for centimeter-level positioning, and a Jetson Xavier NX for onboard computing. A dual-band antenna enables communication with the ground station running QGroundControl with integrated LLM interface. The interface displays a real-time mission map (left) showing the AAV's position and coverage area. A natural language chat window (right) allows the operator to issue high-level commands that the Skypilot converts to flight plans. All experiments were conducted in a 20 m  $\times$ 20 m field partitioned into 100 cells (10  $\times$ 10 grid, 2 m  $\times$ 2 m each). The field comprises four 10 m  $\times$ 10 m quadrants: top-left, top-right, bottom-left, and bottom-right. During missions, the AAV maintains 0.8 m altitude and follows the planner-generated cell sequence. Flight trajectories and LLM outputs are recorded for analysis.

\subsubsection{Trajectory and Key Frame Analysis}

Fig.~\ref{fig:outdoor}(a)--(d) presents four key moments of a single-AAV coverage mission generated by our LLM-based planner. At $t$ = 0 s (Fig.~\ref{fig:outdoor}(a)), the AAV lifts off from the start marker (yellow triangle) and begins a boustrophedon sweep of the bottom-right quadrant while avoiding the virtual  obstacles. By $t$ = 90 s (Fig.~\ref{fig:outdoor}(b)), nearly half of the free  cells  have been visited. The white  path reveals the  Skypilot's quadrant-wise strategy: completing each 10 m $\times$ 10 m quadrant before transitioning to the next, eliminating unnecessary heading reversals.
At $t$ = 150 s (Fig.~\ref{fig:outdoor}(c)), the AAV finishes the lower half and smoothly transitions to the upper quadrants, avoiding cell revisits through lateral shifts. Finally, at $t$ = 200 s (Fig.~\ref{fig:outdoor}(d)), the AAV completes the last unvisited stripe in the top-right corner, achieving 100\% area coverage while respecting all obstacle constraints.

Owing to the heterogeneous demands across four  quadrants, the coverage mission in  Fig.~\ref{fig:outdoor}(e)--(h) is substantially more challenging than a uniform lawn-mower sweep. At $t$ = 0 s (Fig.~\ref{fig:outdoor}(e)), the AAV takes off in the bottom-right quadrant and initiates a boustrophedon sweep while  navigating around a rectangular obstacle located in this normal-density region. By $t$ = 174 s (Fig.~\ref{fig:outdoor}(f)), the AAV  enters the bottom-left quadrant requiring  dense coverage. Here, the trajectory contracts into a maze-like pattern to achieve dense sampling beyond standard cell-center visits.
At $t$ = 288 s (Fig.~\ref{fig:outdoor}(g)), the AAV enters the top-left quadrant designated for rapid transit. The Skypilot produces a direct corridor-style trajectory that threads between the virtual obstacles and exits the quadrant with minimal dwell time before resuming its sweep. Finally, at $t$ = 350 s (Fig.~\ref{fig:outdoor}(h)),
the AAV navigates the top-right quadrant with higher obstacle density, executing a serpentine path through the clutter while maintaining forward momentum. These frames demonstrate Skypilot's ability to: (i) adapt coverage density to task demands, (ii) negotiate varied obstacle fields, and (iii) transition smoothly between quadrants, minimizing mission time and path length.

\section{CONCLUSIONS}
\label{con}
This paper presents Skypilot, an LLM-enhanced framework that integrates MCTS to address hallucination and reproducibility challenges in AAV coverage planning. Our approach combines a diversified action space with physics-informed rewards to ground LLMs in physical constraints while preserving reasoning capabilities. The fine-tuning strategy using 23,000 MCTS samples reduces inference time significantly while preserving planning quality. Both indoor and outdoor experimental results demonstrate Skypilot's superior coverage efficiency and constraint satisfaction compared to baseline methods. Moving forward, this work can be extended to incorporate multi-agent coordination, continual learning, and visual perception in GPS-denied environments.

\bibliographystyle{unsrt}
\bibliography{reference}

\newpage

\vspace{-3em}

\vfill

\end{document}